\documentclass[5p,times]{elsarticle}
\pdfoutput=1 

\usepackage{mathtools}
\usepackage{graphicx}
\usepackage{subcaption}
\usepackage[table,xcdraw]{xcolor}
\usepackage{url}
\usepackage{svg}

\usepackage{amsmath}

\newcommand{\nicehat}[1]{\expandafter\hat#1}   

\usepackage{amssymb}

\journal{Future Generation Computer Systems}

\begin{document}

\begin{frontmatter}



\title{RUAD: unsupervised anomaly detection \protect\\ in HPC systems}

\author[UNIBO]{Martin Molan}

\affiliation[UNIBO]{organization={DISI and DEI Department, University of Bologna},
            city={Bologna},
            country={Italy}}

\author[UNIBO] {Andrea Borghesi}
\author[CINECA]{Daniele Cesarini}
\author[UNIBO,ETH]{Luca Benini}
\author[UNIBO]{Andrea Bartolini}

\affiliation[CINECA]{organization={CINECA consorzio interuniversitario},
            city={Bologna},
            country={Italy}}

\affiliation[ETH]{organization={Institut für Integrierte Systeme, ETH},
            city={Zürich},
            country={Switzerland}}

\begin{abstract}
The increasing complexity of modern high-performance computing (HPC) systems necessitates the introduction of automated and data-driven methodologies to support system administrators' effort toward increasing the system's availability. Anomaly detection is an integral part of improving the availability as it eases the system administrator's burden and reduces the time between an anomaly and its resolution. However, current state-of-the-art (SoA) approaches to anomaly detection are supervised and semi-supervised, so they require a human-labelled dataset with anomalies – this is often impractical to collect in production HPC systems. Unsupervised anomaly detection approaches based on clustering, aimed at alleviating the need for accurate anomaly data, have so far shown poor performance. 

In this work, we overcome these limitations by proposing RUAD, a novel Recurrent Unsupervised Anomaly Detection model. RUAD achieves better results than the current semi-supervised and unsupervised SoA approaches. This is achieved by considering temporal dependencies in the data and including long-short term memory cells in the model architecture. The proposed approach is assessed on a complete ten-month history of a Tier-0 system (Marconi100 from CINECA with 980 nodes). RUAD achieves an area under the curve (AUC) of 0.763 in semi-supervised training and an AUC of 0.767 in unsupervised training, which improves upon the SoA approach that achieves an AUC of 0.747 in semi-supervised training and an AUC of 0.734 in unsupervised training. It also vastly outperforms the current SoA unsupervised anomaly detection approach based on clustering, achieving the AUC of 0.548.

\end{abstract}




\end{frontmatter}


\section{Introduction}
\label{sec:intro}
Recent trends in the development of high-performance computing (HPC) systems (such as heterogeneous architecture and higher-power integration density) have increased the complexity of their management and maintenance \cite{LB1}. A typical contemporary HPC system consists of thousands of interconnected nodes; each node usually contains multiple different accelerators such as graphical processors, FPGAs, and tensor cores \cite{LB2}. Monitoring the health of all those subsystems is an increasingly daunting task for system administrators. To simplify this monitoring task and reduce the time between anomaly insurgency and response by the administrators, automatic anomaly detection systems have been introduced in recent years \cite{LB3}.

Anomalies that result in downtime or unavailability of the system are expensive events. Their cost is primarily associated with the time when the HPC system cannot accept new compute jobs. Since HPC systems are costly and have a limited service lifespan \cite{HPC_trends}, it is in the interest of the system's operator to reduce unavailability times. Anomaly detection helps in this regard as it can significantly reduce the time between the fault and the response by the system administrator, compared to manual reporting of faulty nodes \cite{TPDS}.

Modern supercomputers are endowed with monitoring systems that give the system administrators a holistic view of the system \cite{LB3}. Data collected by these monitoring systems and historical data describing system availability are the basis for Machine Learning anomaly detection approaches \cite{borghesi2019anomaly, borghesi2019frequency, netti2019machine, netti2019online, deep_log}, which build data-driven models of the supercomputer and its computing nodes. In this work, we focus on CINECA Tier0 HPC system (Marconi100 \cite{iannone2018marconi,Marconi100userguide} ranked 9th in Jun. 2020 Top500 list \cite{Top500ref}), which employs a holistic monitoring system called EXAMON \cite{examon_date}.

Production HPC systems are reliable machines that generally have very few downtime events - for instance, in Marconi100 at CINECA,
timestamps corresponding to faulty events represent, on average, only $0.035\%$ of all data. However, although anomalies are rare events, they still significantly impact the system's overall availability - during the observation period, there was at least one active anomaly (unavailable node) $14.4\%$ of the time. State-of-the-art (SoA) methods for anomaly detection on HPC systems are based on supervised and semi-supervised approaches from the Deep Learning (DL) field \cite{TPDS}; for this reason, these methods require a training set with accurately annotated periods of downtime (or anomalies). In turn, this requires the monitoring infrastructure to track downtime events; in some instances, this can be done with specific software tools (e.g., Nagios \cite{barth2008nagios}),  but properly configuring these tools is a complex and time-consuming task for \linebreak system administrators.  

So far, the challenges of anomaly detection on HPC systems have been approached by deploying anomaly reporting tools by training the models in a supervised or semi-supervised fashion \cite{TPDS, martin_exp,tuncer2018online,netti2019machine}. The need for an accurately labelled training set is the main limitation of current approaches as it is expensive, in terms of time and effort of the system administrators, to be applied in practice. Downtime tracking also has to be able to record failures with the same granularity as the other monitoring services. Some methods in production HPC systems only record downtime events by date \cite{LB1,LB2,LB3}. In most production HPC systems, accurate anomaly detection is thus not readily achievable. For this reason, the majority of the methods from the literature were tested on historical or synthetic data or in supercomputers where faults were injected in a carefully controlled fashion \cite{netti2018finj}. Another limitation for the curation of an accurately labeled anomaly dataset is the short lifetime of most HPC systems. In the HPC sector, a given computing node and system technology have a lifetime of between three and five years. Short lifetime means, in practice, that the vendor has no time to create a dataset for training an anomaly detection model before the system is deployed to the customer site. 

A completely unsupervised anomaly detection approach \linebreak could be deployed on a new node or even on an entirely new HPC system. It would then learn online and without any interaction with the system administrators. Additionally, such a system would be easier to deploy as it would require no additional framework to report and record anomalous events (in addition to the monitoring infrastructure needed to build the data-driven model of the target supercomputer - a type of infrastructure which is becoming more and more widespread in current HPC facilities \cite{LB3}).

Unsupervised anomaly detection approaches for HPC systems exist such \linebreak as \cite{k_means_hpc, clustering, anomaly_log}. They either work on log or sensor data. Approaches based on log data \cite{k_means_hpc, anomaly_log}, while useful, can only offer a post-mortem and restricted view of the supercomputer state. The SoA for anomaly detection on sensor data \cite{clustering} is based on clustering, which requires a degree of manual analysis from system administrators and offers poor performance compared to semi-supervised methods. The semi-supervised methods \cite{TPDS, borghesi2019anomaly,borghesi2019online}, based on the dense autoencoders, which are trained to reproduce their input,  could be trained in an unsupervised fashion. However, none of the presented works has explored this possibility. According to the SoA, the models would perform worse as the dense autoencoder is also capable of learning the characteristics of the anomalies \cite{TPDS,borghesi2019anomaly,borghesi2019online}.

The primary motivation for this work is to propose a novel approach that relies \emph{only on the fact that the anomalies are rare events} and works at least equally well when trained in an \emph{unsupervised manner} as it does when trained in \emph{semi-supervised manner} - this has not been the case in the current SoA. In this work, we propose an \emph{unsupervised} approach: RUAD (Recurrent Unsupervised Anomaly Detection) that works on sensor data and \emph{outperforms} all other approaches, including the current SoA semi-supervised approach \cite{TPDS} and SoA unsupervised approach \cite{clustering}. RUAD achieves that by taking into account temporal dependencies in the data. We achieve that by using Long Short-Term Memory (LSTM) cells in the proposed neural network model structure, which explicitly take into consideration the temporal dimension of observed phenomena. We also show that the RUAD model, comprising of LSTM layers, is capable of learning the characteristic of the normal operation even if the anomalous data is present in the test set - the RUAD model is thus able to be trained in an \emph{unsupervised manner}. RUAD targets single HPC computing nodes: we have different anomaly detection models for each computing node. The motivation behind this is \emph{scalability}: in this way, each node can be used to train its own model with minimal overhead - moreover, this strategy would work in larger supercomputers as well, as if the number of nodes increases, we just have to add new detection models.

\subsection{Contributions of the paper}
\label{sec:cont}
To recap, in this paper, we propose an anomaly detection framework that can handle complex system monitoring data, scale to large-scale HPC systems, and be trained even if no labelled dataset is available. The key contributions presented in this paper are:
\begin{itemize}

\item{We propose a completely \emph{unsupervised} anomaly detection approach (RUAD) that exploits the fact that the anomalies are rare and explicitly considers the \emph{temporal dependencies} in the data by using LSTM cells in an autoencoder network. The resulting Deep Learning model \emph{outperforms} the previous state-of-the-art semi-super\-vised \linebreak approach \cite{TPDS}, based on time-unaware autoencoder networks. On the dataset presented and analysed in this paper (collected from the Marconi100 supercomputer), the previous approach achieves an Area-Under-the-Curve (ACU) test set score of $0.7470$. In contrast, our unsupervised approach achieves the best test set AUC score of $0.7672$. To the best of our knowledge, this work is the first time such an approach has been applied to the field of HPC system monitoring and anomaly detection.}



\item{We have conducted a \emph{very large-scale experimental evaluation} of our methods. We have trained four different deep learning models for each of the 980+ nodes of Marconi100. To the best of our knowledge, this is the largest scale experiment relating to anomaly detection in HPC systems, both in terms of the number of considered nodes and length of time. Previous works only evaluate the models on a subset of nodes with a short observation time (\cite{TPDS} paper, for instance, only analyzed 20 nodes of the HPC system over two months). Per-node training of models also demonstrates the feasibility of \emph{per node} models for large HPC systems. The training time for the individual model was under $30$ minutes on a single NVIDIA Volta V100 GPU.}
    
\end{itemize}

\subsection{Structure of the paper}
We present the current state-of-the-art and position our paper in Section \ref{sec:related}. The machine learning approaches used for anomaly detection, including our novel approach, are described in section \ref{sec:meth}. The experimental setting for empirical validation of our results is detailed in Section \ref{sec:setting} and our results are discussed in the rest of Section \ref{sec:ex}. Finally, Section \ref{sec:concl} offers some concluding remarks.

\section{Related Works}
\label{sec:related}
The drive to detect events or instances that deviate from the norm (i.e. \emph{operational anomalies}) is present across many industrial applications. One of the earliest applications of anomaly detection models was credit card fraud detection in the financial industry \cite{anomaly_card, fin_ann_2}. Recently, anomaly detection (and associated predictive maintenance) has become relevant in manufacturing industries \cite{anomaly_industry, industrial_ad}, internet of things (IoT) \cite{IOT_1, IOT_2, IOT_3}, energy sector \cite{anomaly_energy},  medical diagnostics \cite{ad_medicine, medical_2}, IT security \cite{cloud_anomaly}, and even in complex physics experiments \cite{cern}.

Typically, anomalies in an HPC system refer to periods of (and leading to) suboptimal operating modes, faults that lead to failed or incorrectly completed jobs, or node and other components hardware failures. While HPC systems have several possible failure mitigation strategies \cite{f_miti} and fault tolerance \linebreak strategies \cite{f_tol}, anomalies of this type still significantly reduce the amount of compute time available to users \cite{dram_anomaly}. The transition towards Exascale and the increasing heterogeneity of hardware components will only exacerbate the issues stemming from failures, and anomalous conditions that already plague HPC \linebreak machines \cite{LB1, LB3, ex_mon}. A DARPA study estimates that the failures in future exascale HPC systems could occur as frequently as once every 35-39 minutes \cite{DARPA}, thus significantly impacting the supercomputing availability and system administrator load.

However, when looking at specific components and not at the entire HPC system (e.g., considering a single computing node), faults remain very rare events, thus falling under the area of anomaly detection, which can be seen as an extreme case of supervised learning on unbalanced classes \cite{deep_anomaly}. Because data regarding normal operation far exceeds data regarding anomalies, classical supervised learning approaches tend to overfit the normal data and give a sub-optimal performance on the anomalous data \cite{anomaly_rev}. In order to mitigate the problem of unbalanced classes, the anomaly detection problem is typically approached from two angles. Approaches found in the State-of-Art (SoA) that address the class imbalance either modify the data \cite{imblearn} or use specialized techniques that work well on anomaly detection problems \cite{TPDS}. Data manipulation approaches address the dataset imbalance either by decreasing the data belonging to normal operation (\emph{under sampling the majority class}) or by oversampling or even generating anomalous data (\emph{over sampling minority class}) \cite{imblearn}. Data manipulation for anomaly detection in HPC systems has not yet been thoroughly studied. Conversely, most existing approaches rely on synthetic data generation, e.g., injection of anomalies in real (non-production) supercomputers or HPC simulators \cite{TPDS}. 

Another research avenue exploits the abundance of normal data from HPC systems using a different learning strategy,\linebreak, namely semi-supervised ML models. Instead of learning on a dataset containing multiple classes – and consequently learning the characteristics of all classes – semi-supervised models are trained only on the normal data. Hence, they are trained to learn the characteristics of the of the normal class (the majority class in the dataset). Anomalies are then recognized as anything that does not correspond to the learned characteristic of the normal class \cite{deep_anomaly,borghesi2019anomaly,borghesi2019semisupervised,borghesi2019online,glider}. 

Regarding the type of data used to develop and \linebreak deploy anomaly detection systems, we can identify two macro-classes: system monitoring data collected by holistic
monitoring systems (i.e. Examon \cite{examon_date}) and log data. This data is then annotated with information about the system or node-level availability, thus creating a label associated with the data points. The label encodes whether the system is operating normally or experiencing an anomaly. Since it is expensive and time-consuming to obtain labelled system monitoring data, a labelled dataset for supervised learning can be obtained by "injecting" anomalies into the HPC system (like \cite{netti2018finj}). Labels are important for both supervised, semi-supervised and unsupervised approaches. In the first case, they are used to compute the loss, in the second case to identify the training dataset and validation, and in the third case, only for validation. This data can then be used in a supervised learning task directly or after processing new features (feature construction). Examples of this approach are  \cite{tuncer2017diagnosing,tuncer2018online,aksar2021e2ewatch} where authors use supervised ML approaches to classify the performance variations and job-level faults in HPC systems. For fault detection, \cite{netti2019machine,netti2018finj} propose a supervised approach based on Random Forest (an ensemble method based on decision trees) to classify faults in an HPC system. All mentioned approaches use synthetic anomalies injected into the HPC system to train a supervised classification model. Approaches \cite{TPDS} and \cite{martin_exp} are among the few that leverage \emph{real} anomalies collected from production HPC systems (as opposed to injected anomalies). In this paper, we are interested in real anomalies, and thus, we will not include methods using synthetic/simulated data or injected anomalies in our quantitative comparisons. 

All mentioned approaches do not take into account temporal dependencies of data (models are not trained on time series but on \emph{tabular data containing no temporal information}). System monitoring data approach \cite{proctor} is the first to take into account temporal dependencies in data by calculating statistical features on temporal dimension (aggregation, sliding window statistics, lag features).
Most approaches that deal with \emph{time series anomaly detection} do so on system log data. Labelled anomalies are either analyzed with log parsers \cite{inter_HPC} or detected with deep learning methods. Deep learning methods for anomaly detection are based on LSTM  neural networks as they are a proven approach in other text processing fields.

Compared to labelled training sets, much less work has been done on unlabelled datasets - despite this case being much more common in practice. So far, all research on unlabelled datasets has focused on system log data. \cite{k_means_hpc} propose a $k$-means based unsupervised learning approach that does not take into account temporal dynamics of the log data. A clustering-based approach on sensor data is proposed by \cite{clustering}. This approach will serve as one of the baselines in the experimental section (as it is the only unsupervised approach on the sensor and not on log data). An approach \cite{anomaly_log} works on time series data in an unsupervised manner. It uses the LSTM-based autoencoder and is trained on the existing log data dataset. The proposed anomaly detector achieves the AUC (area under the receiver-operator characteristic curve) of $0.59$. Although it works on a drastically different type of dataset (log data as opposed to system monitoring data), it is the closest existing work to the scope of the research presented in this paper. As we show later in the paper, we can achieve much better results than the one reported for the log data models \cite{anomaly_log} by deploying an unsupervised anomaly detection approach on system monitoring data on a per-node basis. Table~\ref{tab:rel} summarizes the most relevant approaches described in this section, focusing on the training set and temporal dependencies. 

\begin{table}[htb]
\begin{center}
\begin{tabular}{|r|c|c|}
\hline
{}          & Tabular data                                                                   & Time series                                            \\ \hline
Supervised &  \multicolumn{1}{c|}{ \cite{e2e, netti2019online}} &  \cite{proctor, inter_HPC, deep_log} \\ \hline

Semi-supervised &  \cite{TPDS, borghesi2019anomaly,borghesi2019semisupervised, borghesi2019online} &  \\ \hline

Unsupervised                   &  \cite{k_means_hpc, clustering}                                          & { \cite{anomaly_log}}        
\\ \hline
\end{tabular}
\caption{Summary of anomaly detection approaches on HPC systems}
\label{tab:rel}
\end{center}
\end{table}

The novelty of this paper is, in relation to the existing works, threefold:
\begin{samepage}
\begin{itemize}
\item{it introduces an
\emph{unsupervised time-series based} anomaly detection model named RUAD;}
\item{it proposes a deep learning architecture that captures \emph{time dependency};}
\item{the approach is evaluated on a \emph{large scale production} dataset with \emph{real anomalies} -- this is the largest scale evaluation ever conducted on this kind of problem, to the best of our knowledge.}
\end{itemize}
\end{samepage}

\section{Methodology}
\label{sec:meth}
In this section, we describe the proposed approach for unsupervised anomaly detection. We do not directly introduce the proposed method (the LSTM autoencoder deep network) as we want to show how it is a significant extension to the current state-of-the-art; thus, we start by introducing three baseline methods, i) exponential smoothing (serving as the most basic method for comparison), ii) unsupervised clustering and iii) the dense autoencoder used in \cite{TPDS}. We then describe our approach in detail and highlight its key strengths (the unsupervised training regime and the explicit inclusion of the temporal dimension).

\subsection{Node anomaly labeling}
\label{sec:lab}
We aim to recognize the severe malfunctioning of a node that prevents it from executing regular compute jobs. This malfunctioning does not necessarily coincide with removing a node for the production, as reported by Nagios. In our discussions with system administrators of CINECA, we have concluded that the best proxy for node availability is the most critical state, as reported by Nagios. For this reason, we have created a \emph{new label} called \emph{node anomaly} that has a value 1 if any subsystem reported by Nagios reports a critical state. From these events (reported anomalies), we then filter out known false positive events based on reporting tests or configurations in Jira \cite{wiki:Jira_(software)}. Jira logs are supplied by CINECA. The labels used in our previous work \cite{TPDS} do not apply to M100 as they were extensively used to denote nodes being removed from production for testing and calibration. In this work, we are examining the early period of the HPC machine life-cycle, when several rounds of re-configuration were performed, thus partially disrupting the normal production flow of the system. Comparing the two labelling strategies in table \ref{tab:labels}, we can see that the overlap between the two is minimal. Additionally, there are far fewer anomalies as reported by the \emph{node anomaly} mainly because the M100 went through substantial testing periods in the first ten months of operation where nodes are marked as removed from production while still functioning normally. In the remainder of the paper, class $0$  or class $1$ will \emph{always} refer to the value of \emph{node anomaly} being $0$ or $1$ respectively. Normal data is all data where \emph{node anomaly} has value 0 and \emph{anomalies} are instances where \emph{node anomaly} has value 1.

\begin{table}[ht]
\resizebox{1\columnwidth}{!}{%
\begin{tabular}{l|cc|}
\cline{2-3}
                                                     & \multicolumn{2}{c|}{Node anomaly}       \\ \cline{2-3} 
                                                     & \multicolumn{1}{c|}{0}          & 1     \\ \hline
\multicolumn{1}{|l|}{Removed from production: False} & \multicolumn{1}{c|}{12 139 560} & 4 280 \\ \hline
\multicolumn{1}{|l|}{Removed form production: True}  & \multicolumn{1}{c|}{15 783}     & 12    \\ \hline
\end{tabular}
}
\caption{Comparison between \emph{removed from production} and \emph{node availability}. The anomalies studied in this work (node availability) significantly differ (and are more reliable) from anomalies studied in previous works. The new labels also mark much fewer events as anomalous.}
\label{tab:labels}
\end{table}

\subsection{Reconstruction error and result evaluation}
\label{sec:def}

The problem of anomaly detection can be formally stated as a problem of training the model $M$ that estimates the probability $P$ that a sequence of vectors of length $W$ ending at time $t_0$ represents an anomaly at time $t_0$:
\begin{equation}\label{eq1}
M: \vec{x}_{t_0-W+1}, \cdots, \vec{x}_{t_0} 	\rightarrow P(\vec{x}_{t_0}\:is\:an\:anomaly).
\end{equation}

Vector $\vec{x_t}$ collects all feature values at time $t$; the features are the sensor measurements collected from the computing nodes. $W$ is the size of the past window that the model $M$ takes as input. If the model does not take past values into account - like the dense model implemented as a baseline \cite{TPDS} - and the window size $W$ is $1$, the problem can be simplified as estimating:
\begin{equation}\label{eq2}
M: \vec{x}_{t_0} 	\rightarrow P(\vec{x}_{t_0}\:is\:an\:anomaly).
\end{equation}

In the case of autoencoders, model $M$ is composed of two parts: autoencoder $A$ (a neural network) and the anomaly score, which is computed using the reconstruction error of the autoencoder. The reconstruction error is calculated by comparing the output of autoencoder model $A$ and the real value vector $\vec{x}_{t_0}$. The task of  model $A$ is to reconstruct the last element of its input sequence:

\begin{equation}\label{eq3}
A: \vec{x}_{t_0-W+1}, \cdots, \vec{x}_{t_0} 	\rightarrow\hat{\vec{x}}_{t_0}.
\end{equation}

Vector $\hat{\vec{x}}_{t_0}$ the reconstruction of vector  $\vec{x}_{t_0}$. As in Eq.~\ref{eq2}, window size $W$ can be $1$. The model $A$ outputs normalized data . The reconstruction error is calculated as the sum of the absolute difference between the output of model $A$ and the normalized input value for each feature: $Error(t_0) = \sum_i^N|\hat{x}_i - x_i|$ where $N$ is the number of features and $\hat{\vec{x}}_{t_0}$ is the output of the model $A$. The error is then normalized by dividing it by the maximum error on the training set: $Normalized\:error(t_0) =  \frac{Error(t_0)}{max(Error(t))}$. We estimate the probability for class 1 (anomaly) as 
\begin{equation}\label{eq4}
P(\vec{x}_{t_0}\:is\:an\:anomaly) =
\begin{cases}
1,\:if:\:Normalized\:error 	\geq 1, \\
Normalized\:error,\:otherwise
\end{cases}
\end{equation}

Based on probability $P(\vec{x}_{t_0}\:is\:an\:anomaly)$, the classifier makes the prediction whether the sequence $\vec{x}_{t_0-W}, \cdots, \vec{x}_{t_0}$ belongs to class $1$ (anomaly) of class $0$ (normal operation). This prediction depends on a threshold $T$, which is a tunable parameter:

\begin{equation}\label{eq5}
Class(\vec{x}_{t_0}) =
\begin{cases}
1,\:if:\:P(\vec{x}_{t_0}\:is\:an\:anomaly)	\geq T, \\
0,\:otherwise
\end{cases}
\end{equation}

To avoid selecting a specific threshold $T$, we introduce the Receiver-Operator Characteristic curve (ROC curve) as a performance metric. It allows us to evaluate the performance of the classification approach for all possible decision thresholds \cite{roc}. The receiver-operator characteristic curve plots the true-positive rate in relation to the false-positive rate. The random decision represents a linear relationship between the two -- for a classifier to make sense, the ROC curve needs to be above the diagonal line. For each specific point on the curve, the better classifier is the one whose ROC curve is above the other. The overall performance of the classifier can be quantitatively computed as the Area Under the ROC Curve (AUC); a classifier making random decisions has the AUC equal to 0.5. AUC scores below 0.5 designate classifiers that are worse than random choice. The best possible AUC score is 1, which is achieved by a classifier that would achieve a true-positive rate equal to 1 while having a false-positive rate equal to 0 (broadly speaking, this is only achievable on trivial datasets or very simple learning tasks).

\subsection{Trivial baseline: exponential smoothing}
Exponential smoothing is implemented as a trivial baseline comparison. It is a simple and computationally inexpensive method that detects rapid changes (jumps) in values. If the anomalies were simply rapid changes in values with no correlation between features, a simple exponential smoothing method would be able to discriminate them. Therefore, we chose exponential smoothing as a first baseline as it is computationally inexpensive and requires no training set.
Additionally, if exponential smoothing performs poorly, this underlines that we are indeed solving a non-trivial anomaly detection problem, for which more powerful models are needed.

For the baseline, we choose to implement exponential smoothing per feature independently. Exponential smoothing for feature $i$ at time $t$ is calculated as:
\begin{equation}\label{eqn:exp}
\hat{x_i} = \alpha x^i_t + (1-\alpha) \expandafter\nicehat{x^{i}_{t-1}},\forall i \in F
\end{equation}

where $\nicehat{x^i}_t$ is an estimate of $x^i$ at time $t$ and $\alpha$ is a parameter of the method. We do this for all features in set $F$. The estimate at the beginning of the observation is equal to the actual value at time $t_0$: $\hat{x}^i_{t_0} = x^i_{t_0}.$

\subsection{Unsupervised baseline: clustering}
A possible approach to unsupervised anomaly detection is to use standard unsupervised machine learning techniques such as k-means clustering proposed by \cite{clustering}. The clusters are determined on the train set; each new instance belonging to the test set is associated with one of the pre-trained clusters.
We opted for this particular unsupervised technique for the comparison as it is the only unsupervised method found in the literature (to the best of our knowledge) which uses sensor data and not logs - and thus, we guarantee a fair comparison. 
It has to be noted, however, that clustering, while belonging to the field of unsupervised machine learning \emph{cannot detect anomalies} in an unsupervised manner - for each of the clusters determined on the train set, the probability for the anomaly has to be calculated. This probability can only be calculated using the labels.

In this work, the clustering approach inspired by \cite{clustering} is implemented to prove the validity of the obtained results. We have used K-means clustering \cite{k_means_hpc} like it has been proposed in \cite{clustering}. We have trained the clusters on the train set. Based on the silhouette score\footnote{the Silhouette score is a measure of performance for a clustering method. It measures how similar an instance is to others in its own cluster compared to instances from the other clusters \cite{silhuete}. It is calculated as $S_{score} = \frac{b-a}{max(a,b)}$ where $a$ is the mean inter-cluster distance, and $b$ is the mean nearest cluster distance for each sample.}  on the train set, we have determined the optimal number of clusters for each node\footnote{Optimal number of clusters is the number of clusters that produces the highest silhouette score on the train set.}. The percentage of instances that belong to class 1 is calculated for each of the determined clusters. We use this percentage of anomalous instances as the anomaly probability for each instance assigned to a specific cluster. The train and test set split is the same as in all other evaluated methods.

\subsection{Semi-supervised baseline: dense autoencoder}
The competitive baseline method is based on the current state-of-the-art dense autoencoder model proposed by \cite{TPDS}. Autoencoders are types of neural networks (NN) trained to reproduce their input. The network is split into two (most often symmetric) parts: encoder and decoder. The role of the encoder is to compress the input into a more condensed representation. This representation is called the \emph{latent layer}. To prevent the network from learning a simple identity function, we choose the latent layer to be smaller than the original input size (number of input features) \cite{borghesi2019anomaly}. The role of the decoder is to reconstruct the original input using the latent representation. 

\begin{figure*}[htb]
    \begin{subfigure}{\textwidth}
        \centering
        \includegraphics[width=\textwidth]{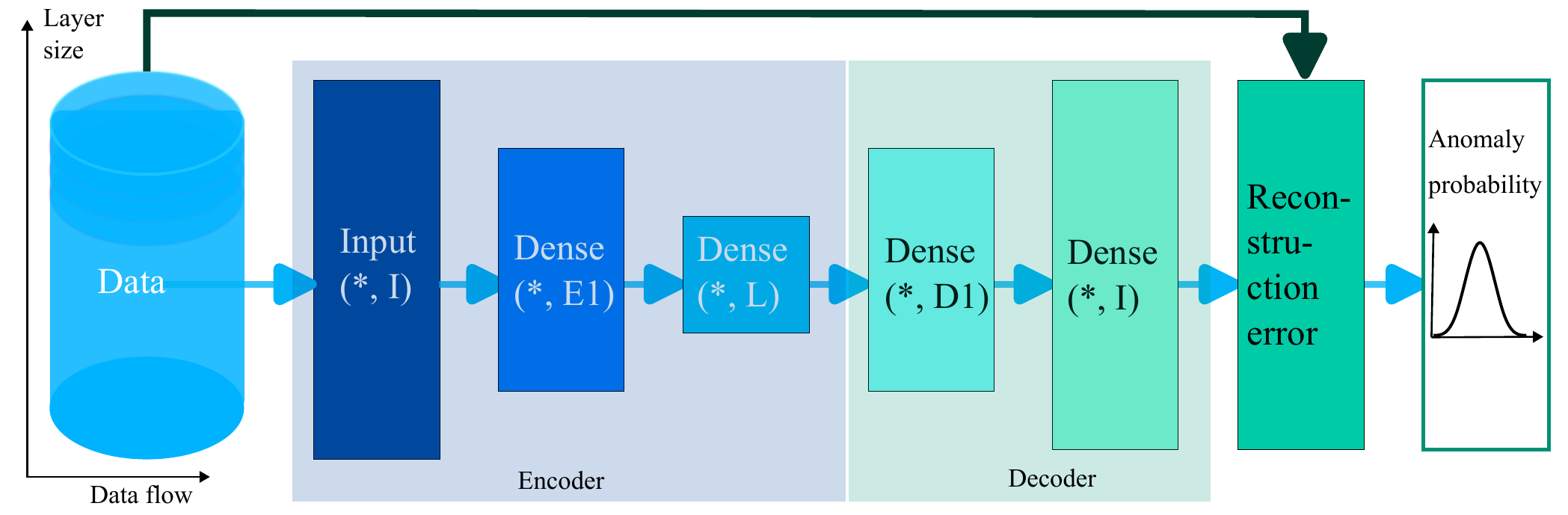}
        \caption{Structure of baseline model - the dense autoencoder.}
        \label{fig:Dense}
    \end{subfigure}
    \begin{subfigure}{\textwidth}
        \centering
        \includegraphics[width=\textwidth]{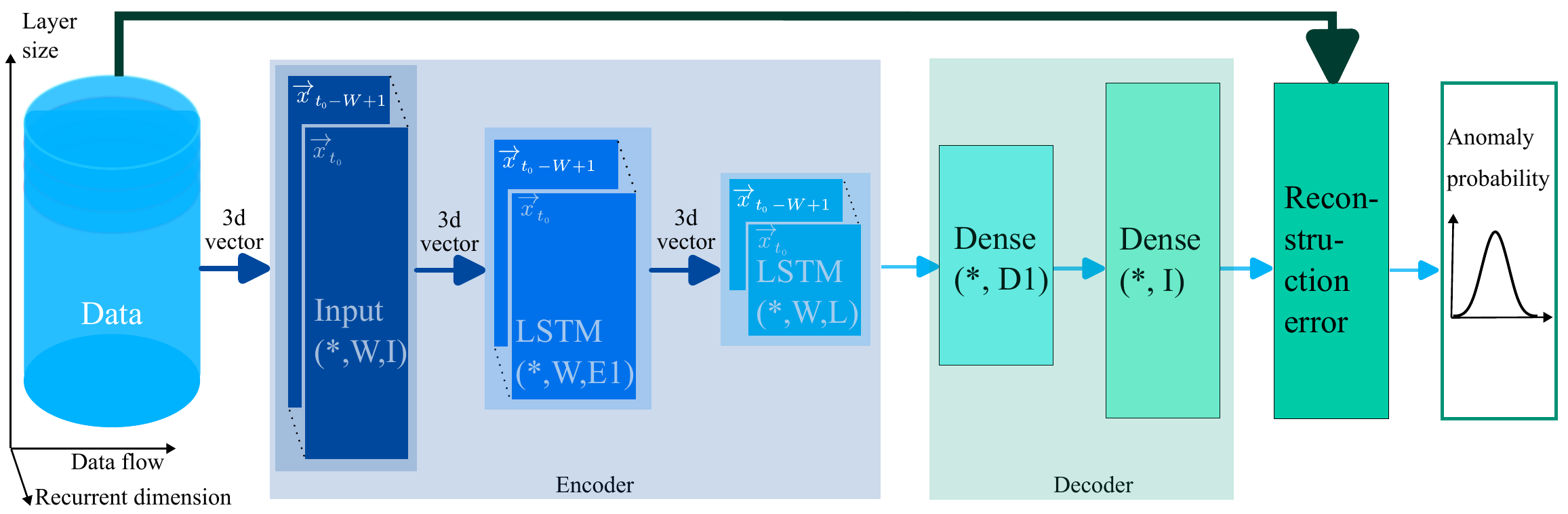}
         \caption{Structure of the proposed RUAD model consisting of the LSTM encoder and dense decoder.}
        \label{fig:LSTM}
    \end{subfigure}
    \caption{The proposed approach replaces the encoder of the baseline model (\ref{fig:Dense}) with the LSTM autoencoder (\ref{fig:LSTM}). The last layer of LSTM encoder returns a vector (not a temporal sequence) which is then passed to the fully connected decoder. $W$ is the window size, $I$ is the size of the input data, $L$ is the size of the latent layer and $E1$ and $D1$ are sizes of encoder and decoder layer respectively. Chosen parameters for $L$, $W$, $E1$ and $D1$ are listed in Section \ref{sec:hyper}.}
    \label{fig:structure}
\end{figure*}

Dense autoencoders are a common choice for anomaly detection since we can restrict their expressive power by acting on the size of the latent layer. Compressing the latent dimension forces the encoder to extract the most salient characteristics from the input data; unless the input data is highly redundant, the autoencoder cannot correctly learn to recreate its input after a certain latent size reduction. In the current state-of-the-art for anomaly detection in production supercomputers (\cite{TPDS}) the dense autoencoder is used in a semi-supervised fashion, meaning that the network is trained using only data points corresponding to the normal operation of the supercomputer nodes (Class 0). Semi-supervised training is doable as the normal points are the vast majority and thus are readily available; however, this requires having labelled data or at least a certainty that the HPC system was operating in normal conditions for a sufficiently long period of time. Once the autoencoder has been trained using only normal data, it will be able to recognize similar but previously unseen points. Conversely, it will struggle to reconstruct new points which do not follow the learned normal behaviour, that is, the anomalies we are looking for; hence, the reconstruction error will be higher. 
The structure of the autoencoder model is presented in Figure \ref{fig:Dense}. The dense autoencoder does not take into account the temporal dynamics of the data -- its input and target output are the same vector:
\begin{equation}\label{eqn:dense}
SoA: \vec{x}_{t_0} 	\rightarrow \vec{x}_{t_0}.
\end{equation}

\subsection{Recurrent unsupervised anomaly detection: RUAD}
\label{sec:RUAD}
Moving beyond the state-of-the-art model, we propose a different approach, RUAD. It takes as input a sequence of vectors and then tries to reconstruct only the last vector in the sequence:

\begin{equation}\label{eqn:lstm}
RUAD: \vec{x}_{t_0-W+1}, \cdots, \vec{x}_{t_0} 	\rightarrow \vec{x}_{t_0}.
\end{equation}

The input sequence length is a tunable parameter that specifies the size of the observation window $W$. The idea of the proposed approach is similar to the dense autoencoder in principle, but with a couple of significant extensions: 1) we are encoding an input sequence into a more efficient representation (latent layer) and 2) we train the autoencoder in an unsupervised fashion (thus removing the requirement of labelled data). The key insight in the first innovation is that while the data describing supercomputing nodes is composed of multi-variate time series, the state-of-the-art does not explicitly consider the temporal dimension -- the dense autoencoder has no notion of time nor of \emph{sequence} of data points. To overcome this limitation, our approach works by encoding the sequence of values \emph{leading up to the anomaly}. The encoder network is composed of Long Short-Term Memory (LSTM) layers, which have been often proved to be well suited to the context where the temporal dimension is relevant \cite{lindemann2021survey}. An LSTM layer consists of recurrent cells that have an input from the previous timestamp and from the long-term memory.


To address the scale of current pre-exascale and future exascale HPC systems that will consist of thousands of nodes \cite{LB3}, we want a scalable anomaly detection approach. The most scalable approach currently for anomaly detection on a whole supercomputer is a node-specific approach as each compute node can train its own model. Still, we want to achieve this by minimally impacting the regular operation of the HPC system. This is why it is important for the proposed solution to have a small overhead. Additionally, since we want to train a per-node model, we want the method to be data-efficient. To address these requirements, we choose not to make the decoder symmetric to the encoder. The proposed approach is thus comprised of a Dense decoder and an LSTM encoder. LSTM encoder output is passed into a dense decoder trained by reproducing the final vector in an input sequence. The decoder network is thus composed of fully connected dense layers. The architecture of the proposed approach is compared to the state-of-the-art approach in Figure \ref{fig:structure}.

The reduced complexity of training allows us to train a separate model for each compute node. As shown previously (\cite{AI_eng}), node-specific models provide better results than a single model trained on all data. We decided to adopt this scheme (one model per node) after a preliminary empirical analysis showed no significant accuracy loss while the training time was vastly reduced (by approximately $50\%$); this is very important in our case as we trained one DL model for each of the nodes of Marconi 100 (980+), definitely a non-negligible computational effort.


\subsection{Data pre-processing}
As introduced in Section \ref{sec:RUAD} our proposed methodology consists of training a model for each node. Thus, the data from each node is first split into training and test sets. The training set contains $80\%$ of data, and the test set contains the last $20\%$ of data (roughly the last two months of data). It is important to stress that we have chosen to have two not overlapping datasets for the training and test. This avoids the cross-transferring of information when dealing with sequencing. Moreover, the causality of the testing is preserved. (No in-the-future data are used to train a model). This makes the results valid for in-practice usage.

\begin{figure*}[ht]
    \centering
    \includegraphics[trim={1cm 8.5cm 1cm 7.5cm},clip, width=1\textwidth]{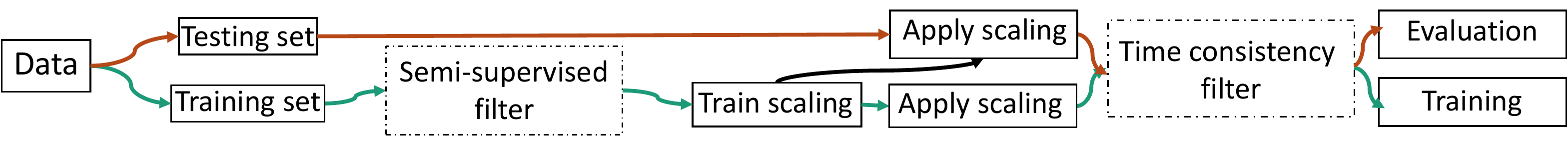}
    \caption{Data processing schema. Data flow is represented by green (training set) and orange (testing set). Scaling is trained on training set and applied on testing set to avoid contaminating the testing set. Semi-supervised and time consistency filters are optional and applied only when required by the modeling approach as indicated in Table \ref{tab:short_names}}
    \label{fig:data_proces}
\end{figure*}

For semi-supervised training, the training set is filtered by removing anomalous events (anomalous events are identified by the \emph{node anomaly} label as described in Section \ref{sec:lab}). We name this filter the semi-supervised filter, as depicted in Figure \ref{fig:data_proces}. For unsupervised learning, the training set is not filtered. For both the cases (unsupervised and semi-supervised learning), labels are used to evaluate the results. After filtering, a scaler is fitted to training data. A scaler is a transformer that scales the data to the $[0,1]$ interval. In the experimental part, a min/max scaler is used on each feature \cite{scikit}.
After fitting to the training data, the scaler is applied to the test data - for rescaling the test set, min and max values of the training set are used (as it is standard practice in DL methods). After scaling, both training and test sets are filtered out to ensure time consistency: the data is split into sequences without missing chunks (missing chunks are the result of the semi-supervised filter). The sequences that are smaller than $W$ are dropped. Finally, sequences are transformed into batches of sequences with length $W$. Figure \ref{fig:data_proces} describes the whole data pre-processing pipeline. 

\subsection{Summary of evaluated methods}

We compare our proposed approach RUAD against established semi-supervised and unsupervised baselines. Summary of pre-processing filters is presented in Table \ref{tab:short_names}. The semi-supervised filter is applied to all semi-supervised approaches. A time consistency filter is applied to methods that explicitly consider the temporal dimension of the data: Exponential smoothing and RUAD. RUAD and the current SoA anomaly detection approach based on dense autoencoders (\cite{TPDS}) is evaluated in both semi-supervised and unsupervised version.

\begin{table}[htb]
\resizebox{1\columnwidth}{!}{
\begin{tabular}{r|cc|c}
\cline{2-3}
\multicolumn{1}{l|}{}                                   & \multicolumn{2}{c|}{Filters}               & \multicolumn{1}{l}{}             \\ \hline
\multicolumn{1}{|r|}{Model}                             & \multicolumn{1}{c|}{Semi-supervised} & Time consistency & \multicolumn{1}{c|}{Name}  \\ \hline

\multicolumn{1}{|r|}{Trivial baseline: exponential smoothing}             & \multicolumn{1}{c|}{NO}              & YES              & \multicolumn{1}{c|}{$EXP$}         \\ \hline
\multicolumn{1}{|r|}{Unsupervised baseline: clustering}                        & \multicolumn{1}{c|}{NO}              & NO               & \multicolumn{1}{c|}{$CLU$}         \\ \hline

\multicolumn{1}{|r|}{DENSE autoencoder baseline semi-supervised} & \multicolumn{1}{c|}{YES}             & NO               & \multicolumn{1}{c|}{$DENSE_{semi}$} \\ \hline
\multicolumn{1}{|r|}{DENSE autoencoder baseline unsupervised}    & \multicolumn{1}{c|}{NO}              & NO               & \multicolumn{1}{c|}{$DENSE_{un}$}   \\ \hline

\multicolumn{1}{|r|}{RUAD semi-supervised}              & \multicolumn{1}{c|}{YES}             & YES              & \multicolumn{1}{c|}{$RUAD_{semi}$}  \\ \hline

\multicolumn{1}{|r|}{RUAD unsupervised}                 & \multicolumn{1}{c|}{NO}              & YES              & \multicolumn{1}{c|}{$RUAD$}    \\ \hline

\end{tabular}
}
\caption{Short names and training strategies for examined methods. $DENSE_{semi}$ is the current SoA \cite{TPDS}.}
\label{tab:short_names}
\end{table}

\begin{table}[htb]
\resizebox{\columnwidth}{!}{
\begin{tabular}{|r|c|c|
}
\hline
Method                           & Training set required & Post-training                             \\ \hline
$EXP$            & Unlabeled dataset     & No action required                        \\ \hline
$CLU$ \cite{clustering}                      & Unlabeled dataset     & Assigning anomaly probability to clusters \\ \hline
$DENSE_{semi}$ \cite{TPDS} & Labeled dataset       & No action required                        \\ \hline
\textbf{$RUAD$}                           & \textbf{Unlabeled dataset}     & \textbf{No action required} \\
\hline
\end{tabular}
}
\caption{Caparison of implemented approaches relating to the training set requirements.}
\label{tab:supervised_comp}
\end{table}

We wish to highlight that, unlike the unsupervised learning baseline \cite{clustering}, our proposed method RUAD requires no additional action after the training of the model. The approach RUAD, proposed in this work, works on an \emph{unlabeled dataset} and requires no additional \emph{post training analysis}. A summary of approaches relating to training set requirements is presented in Table \ref{tab:supervised_comp}.

\section{Experimental results}
\label{sec:ex}

\subsection{Experimental setting}
\label{sec:setting}
The focus of the experimental part of this work is Marconi 100 (M100) HPC system, located in the CINECA supercomputing centre. It is a tier-0 HPC system that consists of 980 compute nodes organized into three rows of racks. Each compute node has 32 core CPU, 256 GB of RAM and 4 Nvidia V100 GPUs. In this work, nodes of the HPC system will be considered independent. This is also in line with the current SoA works \cite{netti2018finj, borghesi2019anomaly, TPDS} where anomaly detection is performed per node. Future works will investigate inter-node dependencies in the anomaly detection task.

The monitoring system in an HPC setting typically consists of hardware monitoring sensors, job status and availability monitoring, and server room sensors. In the case of M100, hardware monitoring is performed by Examon\cite{examon_date}, and system availability is provided by system administrators\cite{barth2008nagios}. This raw information provided by Nagios, however, contains many false-positive anomalies. For this reason, we have constructed a new anomaly label called \emph{node anomaly} described in Section \ref{sec:lab}.

For each of the 980 nodes of M100, a separate dataset was created. Dataset details are explained in Section \ref{sec:dataset}.  $DENSE$ and $RUAD$ models were trained and evaluated on the node-specific training and test sets for each node. The training set consisted of the first eight months of system operation, and the test set comprised the remaining two months. Such testing split ensures a fair evaluation of the model as described in Section \ref{sec:dataset}. For the baseline, the exponential smoothing operation (defined in equation \eqref{eqn:exp}) was applied only over the test set (as the approach requires no training). For each node, the scaler (for min and max scaling) was trained on training data and applied to test data. All results discussed in this section are \emph{combined results from all 980 nodes of M100}.

The dense autoencoder and the RUAD model were trained in two different regimes: semi-supervised and unsupervised. For the semi-supervised training, the semi-supervised filter was applied that removed all data points corresponding to anomalies. In the unsupervised case, no such filtering was performed. It can hence be noticed one of the key advantages of the unsupervised approach: \emph{no data pre-processing needs to be done and no preliminary knowledge about the computing nodes condition is required}.

For all three approaches (exponential smoothing, dense autoencoder and the $RUAD$), the probability for an anomaly (class 1) was estimated from reconstruction error as explained in Section \ref{sec:def}. The probabilities from the test sets of all nodes from a single modelling approach (e.g. RUAD with observation window of length $W = 40$) were collected together to plot the Receiver Operator Characteristic (ROC) curve that is a characteristic for the modelling approach across all nodes. For clustering baseline and exponential smoothing (worst performing baselines), the ROC curve is compared against a dummy classifier which randomly chooses the class.

\subsection{Dataset}
\label{sec:dataset}
\begin{table}[htb]
\centering
\begin{tabular}{|r|c|}
\hline
Source              & Features                                                                                                                                                                                                                                                                                                                                                                                                                                                            \\ \hline
Hardware monitoring & \begin{tabular}[c]{@{}l@{}}ambient temp., dimm{[}0-15{]} temp.,\\ fan{[}0-7{]} speed,  fan disk power,\\ GPU{[}0-3{]} core temp. ,\\ GPU{[}0-3{]} mem temp. ,\\ gv100card{[}0-3{]}, core{[}0-3{]} temp. , \\ p{[}0-1{]} io power, \\ p{[}0-1{]} mem power,\\ p{[}0-1{]} power, p{[}0-1{]} vdd temp. ,\\ part max used,\\ ps{[}0-1{]} input power,\\ ps{[}0-1{]} input voltage,\\ ps{[}0-1{]} output current,\\ ps{[}0-1{]} output voltage, total power\end{tabular} \\ \hline
System monitoring   & \begin{tabular}[c]{@{}l@{}}CPU system, bytes out, CPU idle,\\ proc. run, mem. total,\\ pkts. out, bytes in, boot time,\\ CPU steal, mem. cached, stamp,\\ CPU speed, mem. free, CPU num.,\\ swap total, CPU user, proc. total,\\ pkts. in,  mem. buffers, CPU idle,\\ CPU nice, mem. shared, PCIe, \\ CPU wio, swap free\end{tabular}                                                                                                                                \\ \hline
\end{tabular}
\caption{An anomaly detection model is created only on hardware and application monitoring features. More granular information regarding individual jobs is not collected to ensure the privacy of the HPC system users.}
\label{tab:features}
\end{table}

The dataset used in this work consists of a combination of information recorded by Nagios (the system administration tool used to visually check the health status of the computing nodes) and the Examon monitoring systems; the data encompasses the first ten months of operation of the M100 system. The procedure for obtaining a \emph{node anomaly label} is described in Section \ref{sec:lab}. The features collected in the dataset are listed in table \ref{tab:features}. The data covers 980 compute nodes and five login nodes. Login nodes have the same hardware as the compute nodes but are reserved primarily for job submission and accounting. Thus we removed them from our analysis. The data is collected by the University of Bologna with approval from CINECA\footnote{ CINECA is a public university consortium and the main supercomputing centre in Italy\cite{wiki:CINECA}.}.

In order to align different sampling rates of different reporting services (each of the sensors used has a different sampling frequency), $15$ minute aggregates of data points were created. $15$ minute interval was chosen as it is the native sampling frequency of the Nagios monitoring service (where our labels come from). Four values were calculated for each $15$ minute period and each feature: minimum, maximum, average, and variance.

\subsection{Hyperparameters}
\label{sec:hyper}
Hyper-parameters for all methods discussed in this paper were determined based on initial exploration on the set of $50$ nodes. Chosen parameters performed best on the test from the initial exploration nodes (they achieved the highest AUC score on the test set). Results from the initial exploration set are excluded from the results discussed further in the chapter. Tuned hyperparameters include the structure of the neural nets (number and size of layers) and the smoothing factor of the exponential smoothing:
\begin{itemize}
    \item Exponential smoothing: smoothing factor $\alpha = 0.1$
    \item Clustering: hyper-parameter (number of clusters) is trained on a train set for each node independently.
    \item Dense autoencoder: Structure of the network consists of $5$ layers of shapes:
    (*,462), (*,16), (*,8), (*16), (*462). 
    \item RUAD (LSTM encoder, dense decoder): Structure of the network consists of $5$ layers of shapes: (*,W,462), (*,W,16), (*,W,8), (*,16), (*,462). $W$ is the length of the observation window. Chosen window lengths $W$ were: $5,10,20,40$.
\end{itemize}

\subsection{Exponential smoothing}
\begin{figure}[ht]
    \centering
    \includegraphics[width=0.5\textwidth]{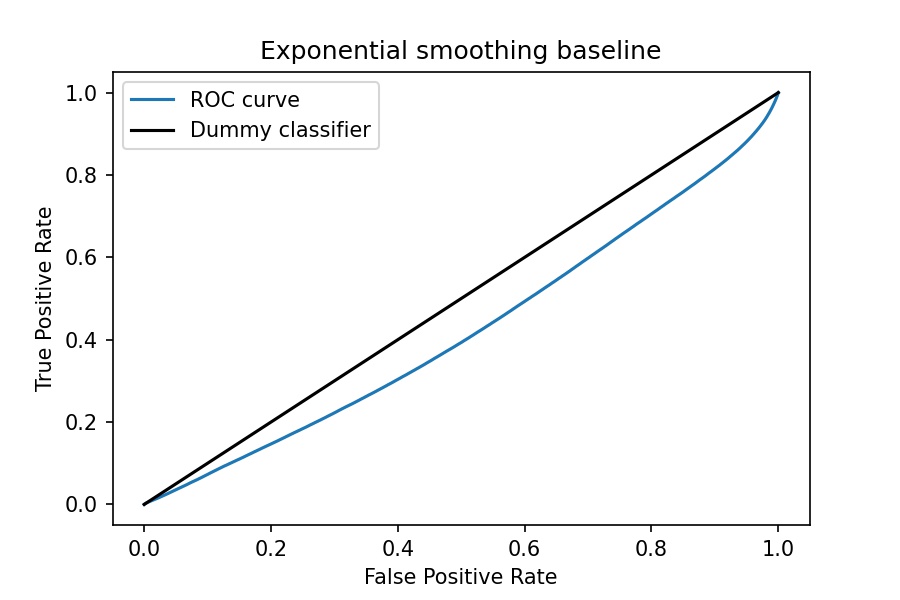}
    \caption{Combined ROC curve from all 980 nodes of M100 for the exponential smoothing baseline. Exponential smoothing performs even worse than the dummy classifier - anomaly detection based on exponential smoothing is completely unusable.}
    \label{fig:EX_results}
\end{figure}

As mentioned in the methodology, exponential smoothing (EXP) is implemented to demonstrate that the anomalies we observe are not simply unexpected spikes in the data signal. Furthermore, exponential smoothing is applied to each feature independently of other features. As shown in Figure \ref{fig:EX_results}, exponential smoothing performs even worse than a dummy classifier (random choice). Poor performance of exponential smoothing shows that the anomalies we are searching for are more complex than simple jumps in values for a feature.

\subsection{Clustering}
\begin{figure}[ht]
    \centering
    \includegraphics[width=0.5\textwidth]{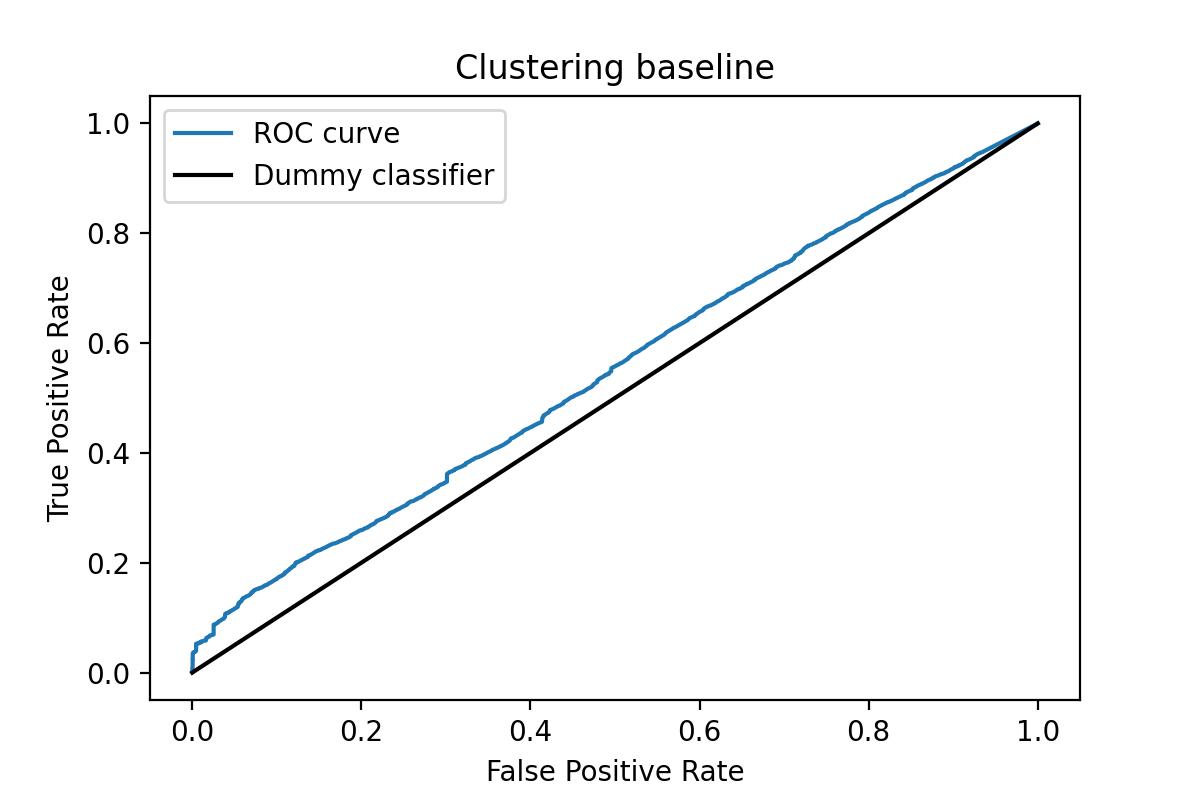}
    \caption{Combined ROC curve from all 980 nodes of M100 for the simple clustering baseline. This baseline performs only marginally better than the dummy classifier.}
    \label{fig:clustering_results}
\end{figure}

The simple clustering baseline performs better than the exponential smoothing baseline and better than the dummy classifier, as seen in Figure \ref{fig:clustering_results}. However, as we will illustrate in the following sections, it performs worse than any other autoencoder method. This demonstrates that the problem we are addressing (anomaly detection on an HPC system) requires more advanced methodologies like semi-supervised and unsupervised autoencoders.

\subsection{Dense autoencoder}

\begin{figure}[htb]
    \centering
    \includegraphics[width=0.5\textwidth]{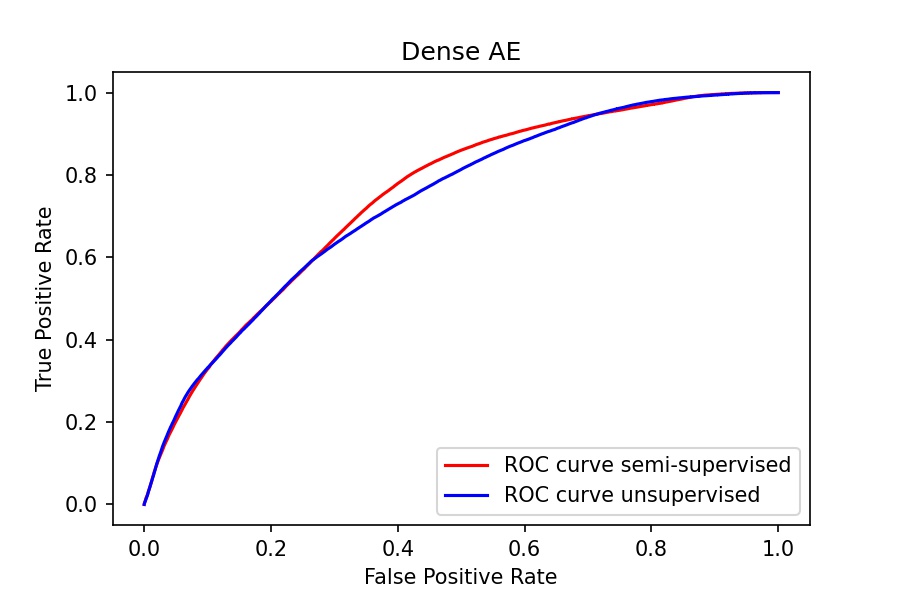}
    \caption{Combined ROC curve from all 980 nodes of M100 for the Dense autoencoder model. In the area interesting for practical application - True Positive Rate between $0.6$ and $0.9$ - semi-supervised approach outperforms unsupervised approach.}
    \label{fig:Dense_results}
\end{figure}

We consider now the dense autoencoder. We train a different network for each computing node of Marconi 100. The optimal network topology was determined during a preliminary exploration done on the sub-sample of the nodes of the system and following the guidelines provided by Borghesi et al.\cite{AI_eng}. In line with the existing work\cite{TPDS}, the semi-supervised learning approach $DENSE_{semi}$ slightly outperforms the unsupervised learning approach $DENSE_{un}$ as seen in Figure \ref{fig:Dense_results}. The better performance in the semi-supervised case is due to the nature of the autoencoder learning model - its capability to reconstruct its input. For example, suppose the autoencoder is fed with anomalous input during the training phase, as in the unsupervised case. In that case,  anomalous examples in the training data constitute a type of ``noise'' that renders the autoencoder partially capable of reconstructing the anomalous examples in the test set.

\subsection{RUAD}
\begin{figure*}[htb]
\centering
\subfloat[Window length 5]{
  \includegraphics[width=0.5\textwidth]{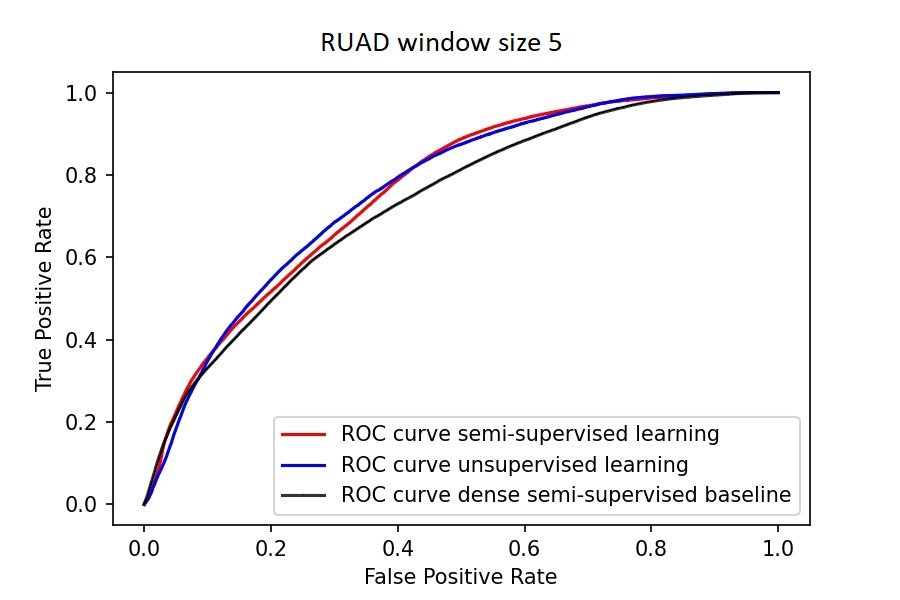}
}
\subfloat[Window length 10]{
  \includegraphics[width=0.5\textwidth]{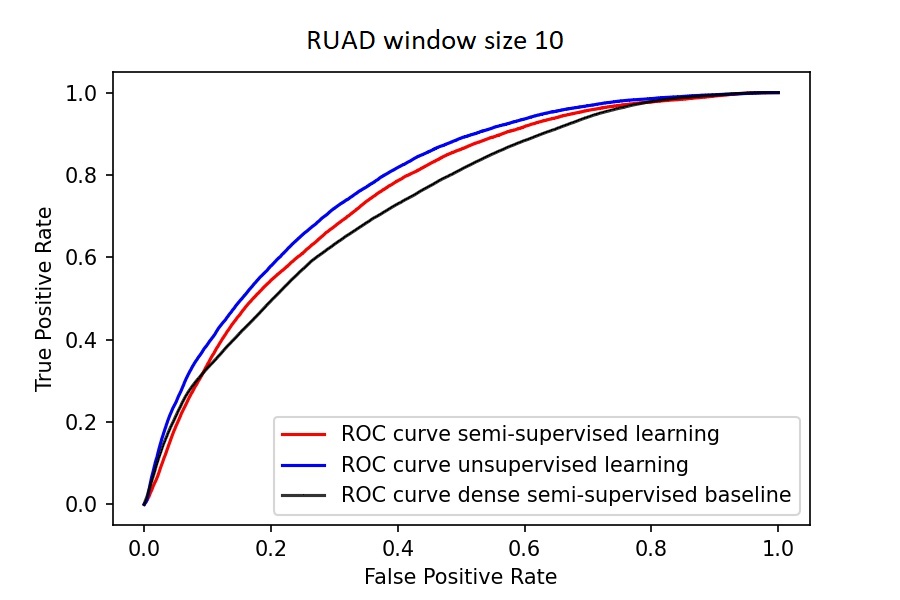}
}
\newline
\subfloat[Window length 20]{
  \includegraphics[width=0.5\textwidth]{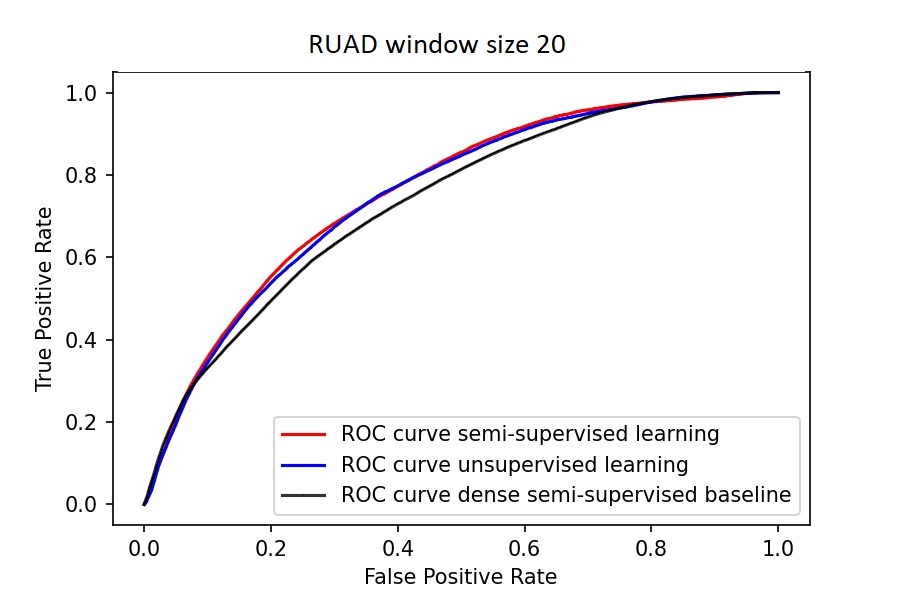}
}
\subfloat[Window length 40]{
  \includegraphics[width=0.5\textwidth]{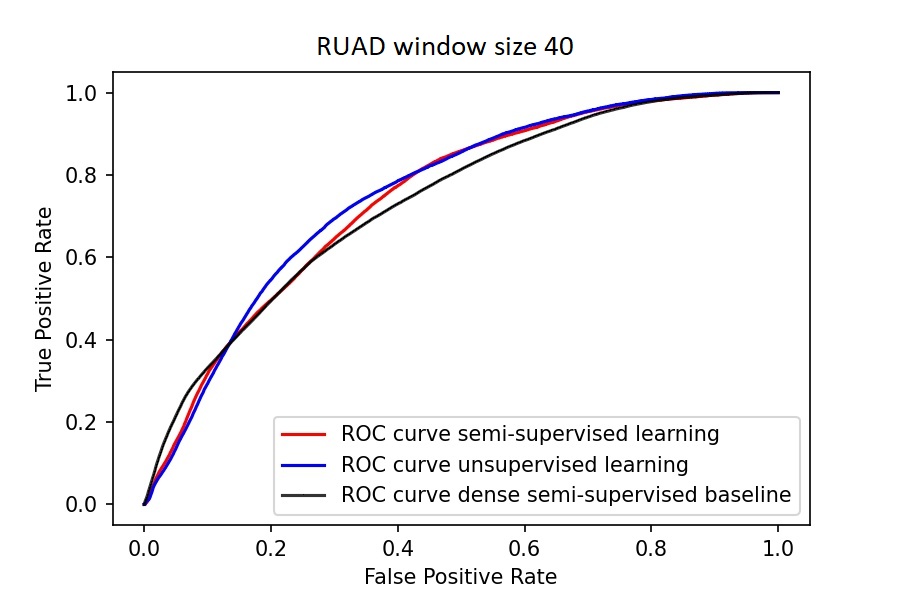}
}
\newline
\hbox to 67.5mm{}
    
\caption{Combined results from all 980 nodes of M100. Comparison of different window lengths for the RUAD model. For all window lengths, performances of semi-supervised and unsupervised approaches are similar. Performance of the proposed model (red and blue line) is compared to the state-of-the-art baseline semi-supervised autoencoder proposed by Borghesi et al.\cite{TPDS}.}
\label{fig:LSTM_res}
\end{figure*}

This section examines the experimental results obtained with the RUAD model (unsupervised LSTM autoencoder). The most important parameter is the length of the input sequence $W$ that is passed to the model. This parameter encodes our expectation of the length of the dependencies within the data. Since each data point represents 15 minutes of node operation, the actual period we observe consists of $W\times15\:min$. In this set of experiments, we selected the following time window sizes: 5 (75 minutes), 10 (2h30), 20 (5h), 40 (10h). These period lengths were obtained after a preliminary empirical evaluation; moreover, these time frames are in line with the typical duration of HPC workloads, which tend to span between dozens of minutes to a few hours\cite{calzarossa2016workload}. We have trained the model in both semi-supervised $RUAD_{semi}$ and unsupervised $RUAD$ fashion for each selected window length. Results across all the nodes are collected in Figure \ref{fig:LSTM_res}.

\subsection{Comparison of all approaches}
The main metric for evaluating model performance is the area under the ROC curve (AUC). This metric estimates the classifiers' overall performance without the need to set a discrimination threshold \cite{roc}. The closer the AUC value is to 1, the better the classifier performs. AUC scores for implemented methods are collected in table \ref{tab:all_res}. From the lower table in table \ref{tab:all_res} (rows correspond to different training regimes and columns to window size for $RUAD$ network) and upper table in \ref{tab:all_res} (rows correspond to the performance of different implemented baselines), we see that the proposed approach outperforms the existing baselines. The highest AUC achieved by the previous baselines is $0.7470$ (achieved by the $DENSE_{semi}$. This is outperformed by $RUAD$ for \emph{all window sizes}. The best performance of $RUAD$ is achieved by selecting the windows size 10 where it achieves an AUC of 7.672. This result clearly shows that some temporal dynamics contribute to the appearance of anomalies.

The final consideration is the impact of observation window length $W$ on the performance of the RUAD model. One might expect that considering longer time sequences would bring benefits, as more information is provided to the model to recreate the time series. This is, however, not the case (as seen in table \ref{tab:all_res}) as the $RUAD$ achieves the best performance of 0.7672 with window size 10. The performance then reduces sharply with window size 40, only achieving an AUC of 0.7473. Several factors might explain this phenomenon. For instance, in tens of hours, the workload on a given node might change drastically. Considering longer time series might thus force the RUAD model to concentrate on multiple workloads, hindering its learning task. 
Finally, an issue stems from the fact that there are gaps (periods of missing measurements) in the collected data (a very likely problem in many real-world scenarios). Longer sequences mean that more data has to be cut from the training set to ensure time-consistent sequences; this is because we are not applying gap-filling techniques at the moment\footnote{We decided not to consider such techniques for the moment, as we wanted to focus on the modelling approach and gap-filling methods tend to require additional assumptions and to introduce noise in the data.}, thus, sub-sequences missing some points need to be removed from the data set. Combining these two factors contributes to the model's decline in performance with longer observation periods.

Considering all discussed factors, the optimal approach is to use the proposed model architecture with window size $W=10$ (i.e. 2 hours and 30 minutes), trained in an \emph{unsupervised} manner. This configuration outperforms semi-supervised $RUAD_{semi}$ as well as the dense autoencoder. As mentioned in the related work (Section \ref{sec:related}), labelled datasets are expensive to obtain in the HPC setting. Good unsupervised performance is why this result is promising - it shows us that if the anomalies represented a small fraction of all data, we could train an anomaly detection model even on an unlabeled dataset (in an unsupervised manner). Such a model not only achieves the state-of-the-art performance but \emph{outperforms} semi-supervised approaches. The best AUC, achieved by the previous SoA $DENSE_{semi}$, is 0.7470. The best AUC score achieved by $RUAD$ is 0.7672. Moreover, unsupervised training makes this anomaly detection model more applicable to a typical HPC (or even datacentre) system. 

\begin{table}[htb]
\centering
\resizebox{0.5\columnwidth}{!}{%
\begin{tabular}{|r|c|}
\hline

Method                          & Combined ROC score \\ \hline
$EXP$                          & 0.4276 \\ \hline
$CLU$    & 0.5478 \\ \hline
$DENSE_{semi}$ & 0.7470 \\ \hline
$DENSE_{un}$    & 0.7344 \\ \hline
\end{tabular}
}

\vspace*{0.2 cm}

\centering
\resizebox{\columnwidth}{!}{%
\begin{tabular}{|r|cccc|}
\hline
Method             & \multicolumn{4}{c|}{Combined ROC score}                                                                                                                                                                      \\ \hline
\rowcolor[HTML]{FFFFFF} 
Sequence length    & \multicolumn{1}{c|}{\cellcolor[HTML]{FFFFFF}5}               & \multicolumn{1}{c|}{\cellcolor[HTML]{FFFFFF}10}              & \multicolumn{1}{c|}{\cellcolor[HTML]{FFFFFF}20}              & 40              \\ \hline
\rowcolor[HTML]{FFFFFF} 
$RUAD_{semi}$ & \multicolumn{1}{c|}{\cellcolor[HTML]{FFFFFF}0.7632}          & \multicolumn{1}{c|}{\cellcolor[HTML]{FFFFFF}0.7582}          & \multicolumn{1}{c|}{\cellcolor[HTML]{FFFFFF}0.7602}          & 0.7446          \\ \hline
\rowcolor[HTML]{FFFFFF} 
$RUAD$      & \multicolumn{1}{c|}{\cellcolor[HTML]{FFFFFF}\textbf{0.7651}} & \multicolumn{1}{c|}{\cellcolor[HTML]{FFFFFF}\textbf{0.7672}} & \multicolumn{1}{c|}{\cellcolor[HTML]{FFFFFF}\textbf{0.7655}} & \textbf{0.7473} \\ \hline
\end{tabular}
}
\vspace*{0.3 cm}
\caption{According to expectations, the semi-supervised dense autoencoder outperforms the unsupervised dense one (highlighted by the higher AUC score). $RUAD$ and $RUAD_{semi}$ \emph{outperform all previous baselines}. In contrast to the dense autoencoders, the proposed approach $RUAD$ performs best in \emph{unsupervised manner}.}
\label{tab:all_res}
\end{table}

\section{Conclusions}
\label{sec:concl}
The paper presents an anomaly detection approach for HPC systems (RUAD) that outperforms the current state-of-the-art approach based on the dense autoencoders \cite{TPDS}. Improving upon state-of-the-art is achieved by deploying a neural network architecture that considers the temporal dependencies within the data.
The proposed model architecture achieves the highest AUC of $0.77$ compared to $0.75$, which is the highest AUC \linebreak achieved by the dense autoencoders (on our dataset).

Another contribution of this paper is that the \linebreak proposed method – unlike the previous work \cite{TPDS, martin_exp,tuncer2018online,netti2019machine} – achieves the best results in an \emph{unsupervised training} case. Unsupervised training is instrumental as it offers a possibility of deploying an anomaly detection model to the cases where (accurately) labelled dataset is unavailable. The only stipulation for the deployment of \emph{unsupervised} anomaly detection models is that the anomalies are rare – in our work, the anomalies accounted for only $0.035\%$ of the data. The necessity to have a few anomalies in the training set, however, is not a significant limitation as HPC systems are already highly reliable machines with low anomaly rates \cite{fugaku, LB1}. 

To illustrate the capabilities of the approach proposed in this work, we have collected an extensive and accurately labelled dataset describing the first $10$ months of operation of the Marconi100 system in CINECA \cite{wiki:CINECA}. The creation of \emph{accurately labelled} dataset was necessary to compare the performance of different models on the data rigorously. Because of the high quality and large scale of the available dataset, we can conclude that for the model proposed in the paper, the unsupervised model \emph{outperforms} semi-supervised model \emph{even if accurate anomaly labels are available}. This is the first experiment of this type and magnitude conducted on a real, in-production datacentre (both in terms of the number of computing nodes considered and the length of the observation period). 

In future works, we will further explore the problem of anomaly detection in HPC systems, in particular, discovering the root causes of the anomalies - e.g., \emph{why} a computing node is entering a failure state? We also have plans to further extend and refine the collected dataset and make it available to the public, in accordance with the facility owners and regulations about users' personal data (albeit not considered in this work, information about the users submitting the jobs to the HPC system can indeed be collected). Moreover, in this work, we focused on node-level anomalies; this was done to be comparable with the state-of-the-art and for scalability purposes; in the future, we will explore the possibility of detecting systemic anomalies as well, i.e., anomalies involving multiple nodes at the same time. In this direction, the natural follow-up to the present work is to build hierarchical approaches which generate anomaly signals based on the composition of the signals generated by the node-specific detection models. 
%

\section{Acknowledgments}
\label{ack}
This research was partly supported by the EuroHPC EU PILOT project (g.a. 1010\-34126), the EuroHPC EU Regale project (g.a. 956560), EU H2020-ICT-11-2018-2019 IoTwins project \linebreak (g.a. 857191), and EU Pilot for exascale EuroHPC EUPEX (g. a. 101033975). We also thank CI\-NE\-CA for the coll\-abo\-ration and access to their machines and Francesco Beneventi for maintaining Examon.






 \bibliographystyle{elsarticle-num} 
 \bibliography{main}





\end{document}